\documentclass[final]{cvpr}
\usepackage{makecell}
\usepackage{times}
\usepackage{epsfig}
\usepackage{graphicx}
\usepackage{amsmath}
\usepackage{amssymb}
\usepackage{multicol}
\usepackage{multirow}
\newcommand{\tablestyle}[2]{\setlength{\tabcolsep}{#1}\renewcommand{\arraystretch}{#2}\centering\small}
\newlength\savewidth\newcommand\shline{\noalign{\global\savewidth\arrayrulewidth
  \global\arrayrulewidth 1pt}\hline\noalign{\global\arrayrulewidth\savewidth}}

\usepackage[pagebackref=true,breaklinks=true,colorlinks,bookmarks=false]{hyperref}

\def\cvprPaperID{10} 
\def\confYear{CVPR 2021}

\begin{document}

\title{Exploring Stronger Feature for Temporal Action Localization}

\author{Zhiwu Qing$^{1,2}$ \quad Xiang Wang$^{1,2}$ \quad Ziyuan Huang$^{2}$\quad Yutong Feng$^2$ \quad Shiwei Zhang$^{2*}$ \\ Jianwen Jiang$^2$  \quad Mingqian Tang$^2$
\quad  Changxin Gao$^1$ \quad Nong Sang$^{1*}$
\\
$^1$Key Laboratory of Image Processing and Intelligent Control \\ School of Artificial Intelligence and Automation, Huazhong University of Science and Technology\\
$^2$Alibaba Group\\
{\tt\small \{qzw, wxiang, cgao, nsang\}@hust.edu.cn}\\
{\tt\small \{pishi.hzy, yutong.fyt, zhangjin.zsw, jianwen.jjw, mingqian.tmq\}@alibaba-inc.com}
}

\maketitle

\let\thefootnote\relax\footnotetext{$*$ Corresponding authors.}
\let\thefootnote\relax\footnotetext{This work is supported by Alibaba Group through Alibaba Research Intern Program.}
\let\thefootnote\relax\footnotetext{This work is done when Z. Qing and X. Wang (Huazhong University of Science and Technology), Z. Huang (National University of Singapore) and Y. Feng (Tsinghua University) are interns at Alibaba Group.}

\begin{abstract}
   Temporal action localization aims to localize starting and ending time with action category. Limited by GPU memory, mainstream methods pre-extract features for each video. Therefore, feature quality determines the upper bound of detection performance. In this technical report, we explored classic convolution-based backbones and the recent surge of transformer-based backbones. We found that the transformer-based methods can achieve better classification performance than convolution-based, but they cannot generate accuracy action proposals. In addition, extracting features with larger frame resolution to reduce the loss of spatial information can also effectively improve the performance of temporal action localization. Finally, we achieve 42.42\% in terms of mAP on validation set with a single SlowFast~\cite{feichtenhofer2019slowfast} feature by a simple combination: BMN~\cite{lin2019bmn}+TCANet~\cite{qing2021tca}, which is 1.87\% higher than the result of 2020~\cite{qing2020tfn}'s multi-model ensemble. Finally, we achieve \textbf{Rank 1st} on the CVPR2021 HACS supervised Temporal Action Localization Challenge.
  
\end{abstract}

\section{Introduction}
 Temporal action localization is a challenging task, especially for HACS dataset~\cite{zhao2019hacs}, which contains complex relationships between actors and scenes in long videos. 
 In this technical paper, we explore two kinds of backbones, \emph{i.e.}, Transformer-based ViViT~\cite{arnab2021vivit} and Timesformer~\cite{bertasius2021timesformer}, CNN-based SlowFast~\cite{feichtenhofer2019slowfast} and CSN~\cite{tran2019csn}.
 From the experiment results, we draw several following conclusions:
 1) The features extracted by the network with remarkable classification performance may not necessarily generate high-quality proposals for temporal action localization. Since the action classification task does not have to be sensitive to the background. For instance, the action that occurs on the football field is likely to play football. The network may only focus on the football field, and there is no need for the act of playing football.
 2) Most videos are rectangular rather than square. When training the network, the input video frames to network are always square. If the same shape is still employed in extracting features, the spatial content will be lost in the rectangular video frames, which is crucial for temporal action detection. 
 %

 %

\begin{figure*}
\begin{center}
\includegraphics[width=15cm]{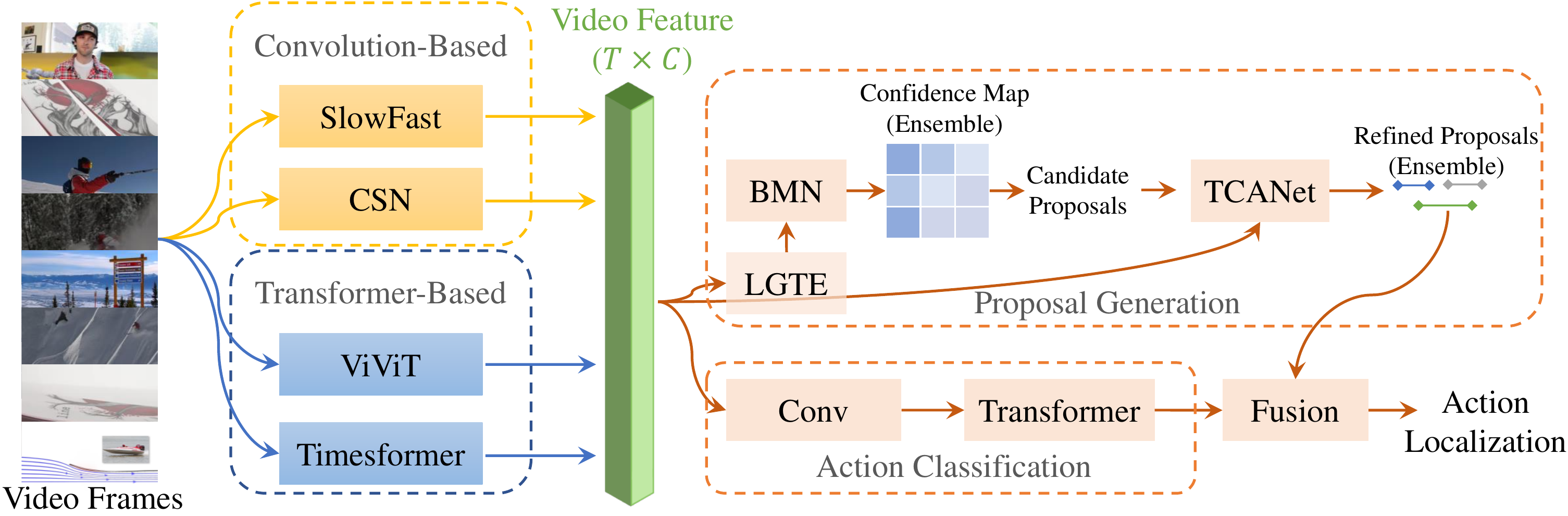}
\end{center}
   \caption{\textbf{The overall framework of our approach.} The input video frames are first extracted features by backbones. Then the BMN~\cite{lin2019bmn} and TCANet~\cite{qing2021tca} are employed to generate accuracy proposals. The video-level features are utilized to perform action classification task. The fusion of proposals and action category generates action detection results.}
\label{fig:framework}
\end{figure*}

\section{Our Approach}
The overall architecture of our approach is visualized in Figure~\ref{fig:framework}. The video features are first extracted from video frames by convolution-based and transformer-based backbones. Then the video features are employed to generate proposals and classify action categories. Finally, the action localization results are generated by fusing proposals with classification scores.

\subsection{Training Backbones}
The existing mainstream pre-training methods can be divided into two types: supervised~\cite{tran2019csn,feichtenhofer2019slowfast,arnab2021vivit,carreira2017i3d} and unsupervised~\cite{huang2021mosi,han2020coclr}. Supervised methods can achieve stronger performance, but need to provide labels for each video. Unsupervised methods can make full use of unlabeled data. We utilize the supervised strategy for pre-training to achieve better performance.

All backbones we employed are first pre-trained on the large-scale Kinetics-700~\cite{carreira2019k700} dataset or Kinetics-600~\cite{carreira2018kinetics600} dataset to improve the generalization ability, and then fine-tuned on the HACS~\cite{zhao2019hacs} dataset.
We explored four backbones with different architectures. As shown in Figure~\ref{fig:framework}, the SlowFast~\cite{feichtenhofer2019slowfast} and CSN~\cite{tran2019csn} are based on convolution, and ViViT~\cite{arnab2021vivit} and Timesformer~\cite{bertasius2021timesformer} are based on transformer. In fine-tuning stage, the features extracted by backbone are used to perform the classification task with $(C+1)$ categories. The fine-tuning details are introduced in the Table~\ref{tab:training-details}. Note that there are differences in training settings between different models. Our training strategies are for reference only. Among them, we use open source pre-trained models to initialize SlowFast, CSN, and Timesformer, and only ViViT is reprodced by us. The details of pre-training process can be referred to our EPIC-KITCHENS-100 Action Recognition report~\cite{huang2021epic}.

\begin{table}[t]
    \centering
\tablestyle{4pt}{1.0}
\small
\begin{tabular}{c|cccc}
     Backbone & \makecell{SlowFast \\ ~\cite{feichtenhofer2019slowfast}} &  \makecell{CSN\\~\cite{tran2019csn}} &  \makecell{ViViT\\~\cite{arnab2021vivit}} &  \makecell{Timesformer\\~\cite{bertasius2021timesformer}}\\
    \shline
    \makecell[c]{Layers}  & 101 & 152  & 16 & 12 \\
    \makecell[c]{Frames}  & 32$\times$2 & 32$\times$2 & 32$\times$2 & 8$\times$32 \\
    \makecell[c]{Resolution}  & \multicolumn{4}{c}{224$\times$224} \\
    \makecell[c]{BS}  & \multicolumn{4}{c}{256} \\
    \makecell[c]{Optimizer}  & SGD & AdamW  & AdamW & SGD \\
    \makecell[c]{LR}  & 0.02 & 1e-4  & 1e-5 & 0.004 \\
    \makecell[c]{WD}  & 1e-7 & 1e-4  & 0.1 & 1e-4 \\
    \makecell[c]{LR Policy}  & \multicolumn{4}{c}{Cosine} \\
    \makecell[c]{WU Epochs}  & 5 & 8  & 2.5 & 1 \\
    \makecell[c]{T Epochs}  & 50 & 30  & 30 & 15 \\
    \makecell[c]{Dropout}  & 0.5 & 0.0  & 0.5 & 0.5 \\

   
\end{tabular}\\
    \caption{\textbf{Training details for all backbones.} ``LR" refers to Learning Rate, ``WD" is weight decay, ``WU" means Warm Up, ``T Epochs" is Training Epochs and ``BS" refers to Batch Size.}
    \label{tab:training-details}
\end{table}

\subsection{Extracting Features}
Following mainstream action proposal generation methods~\cite{lin2019bmn, lin2018bsn,lin2020dbg, xu2020gtad,gao2017turn,buch2017sst,bai2020bcgnn,zhang2019glnet,qing2021tca,qing2020tfn,wang2020cbr,wang2021self}, we pre-extract features for each video. Specifically, for a video which contains $l$ frames, the whole video can be divided into $N$ clips uniformly. We set the stride between consecutive clips to $\delta=8$, which can be converted to 0.267s in a 30-fps videos. In the temporal dimension, the sampling strategy for input clips is consistent with the fine-tuning process. In the spatial dimension, the transformer-based methods are also consistent with fine-tuning, while the convolution-based methods utilize the resolution of 256$\times$320 as the input to extract features, which can save more spatial information for temporal localization.

\subsection{Generating Proposals}

The popular Boundary Matching Network(BMN)~\cite{lin2019bmn} based on dense prediction are easier to generate proposals with high recall rate. Combined with Temporal Context Aggregation Network(TCANet)~\cite{qing2021tca} to further refine proposals, it can achieve impressive performance on HACS dataset. 

\textbf{Training BMN.} The video-level features($C\times N$) are resized to 200, (e.g. $C\times 200$). Local-Global Temporal Encoders(LGTEs)~\cite{qing2021tca} are also inserted into the base module in the BMN. The AdamW~\cite{loshchilov2017adamw} is employed as optimizer. The batch size, learning rate, weight decay and training epochs are set to 128, 0.001, 1e-5 and 10, respectively. BMN designs Temporal Evaluation Module(TEM) and Proposal Evaluation Module(PEM) to evaluate the boundary scores and the IoU of proposals. In our implementation, we only employ the scores output by PEM, since the output of TEM lack global perceptions, which cannot improve the precision.

\textbf{Training TCANet.} We do not resize the features to preserve fine-grained temporal information. Three Temporal Boundary Regressors(TBRs)~\cite{qing2021tca} are employed to refine the proposals generated by BMN, and the first TBR is employed to augment proposals~\cite{wan2021adpi} for accurate proposal distribution. Our optimizer for TCANet is Adam~\cite{kingma2014adam}, and the batch size, learning rate and weight decay is set to 64, 0.0016 and 1e-5, respectively. We train the models for 10 epochs with cosine learning rate schedule.

\textbf{Suppressing redundant predictions.} We utilize Soft-NMS~\cite{bodla2017softnms} to remove redundant predictions. The low threshold, high threshold and alpha in Soft-NMS are set to 0.25, 0.9 and 0.4, respectively.

\subsection{Generating Detection Results}
Since the proposals output by BMN~\cite{lin2019bmn} and TCANet~\cite{qing2021tca} are class-agnostic, they need to be further classified to generate detection results. Considering that almost all videos in the HACS dataset~\cite{zhao2019hacs} have only one category, we directly fuse the video-level classification results with proposals:
\begin{equation}
    S_{det} = S_{props} \times S_{action}.
\end{equation}
 Where the $S_{det}$ is the final detection score for submission, the $S_{props}$ is the score for each proposal output by BMN or TCANet, and the $S_{action}$ is the video-level score for each category.

\begin{table}[t]
    \centering
\tablestyle{4pt}{1.0}
\small
\begin{tabular}{c|c}
    Backbone  & Top-1(Val)\\
    \shline
    CSN~\cite{tran2019csn}& 91.54\% \\
    SlowFast~\cite{feichtenhofer2019slowfast} & 90.37\% \\
    ViViT~\cite{arnab2021vivit} &  \textbf{91.92}\%\\
    Timesformer~\cite{bertasius2021timesformer}& 91.81\%\\

\end{tabular}\\
    \caption{\textbf{Comparison between different backbones for clip-level action classification.} 
    }
    \label{tab:cls}
\end{table}

\begin{table}[t]
    \centering
\tablestyle{4pt}{1.0}
\small
\begin{tabular}{c|c|c}
    Feature & LGTE~\cite{qing2021tca} & mAP(Val)\\
    \shline
    CSN~\cite{tran2019csn} & - & 38.88\% \\
    CSN~\cite{tran2019csn} & x2 & \textbf{40.88\%} \\
    SlowFast~\cite{feichtenhofer2019slowfast} & - & 38.83\% \\
    SlowFast~\cite{feichtenhofer2019slowfast} & x2 & 39.77\% \\
    ViViT~\cite{arnab2021vivit} & - & 36.63\%\\
    ViViT~\cite{arnab2021vivit} & x2 & 37.30\%\\
    Timesformer~\cite{bertasius2021timesformer}& - & 32.23\%\\

\end{tabular}\\
    \caption{\textbf{Comparison between different features based on BMN.} The ``x2" means that we insert 2 LGTEs into base module in BMN.
    }
    \label{tab:feature_results}
\end{table}

\begin{table}[t]
    \centering
\tablestyle{4pt}{1.0}
\small
\begin{tabular}{c|cc|c}
    Feature & Resolution & LGTE~\cite{qing2021tca} & mAP(Val)\\
    \shline
    \multirow{4}{*}{SlowFast\cite{feichtenhofer2019slowfast}}& 224x224 & - & 37.28\%\\
    ~& 224x224 & x2 & 38.39\%\\
    ~& 256x320 & -  & 38.83\%\\
    ~& 256x320 & x2  & \textbf{39.91\%} \\
\end{tabular}\\
    \caption{\textbf{Ablation studies for feature resolution.} The resolution refers to the input resolution of frames when extracting feature.}
    \label{tab:resolution_results}
\end{table}

\begin{table}[t]
    \centering
\tablestyle{4pt}{1.0}
\small
\begin{tabular}{cc|c|cc}
    Feature & Method & Top-1(Val) & \makecell[c]{mAP\\(val)} & \makecell[c]{mAP\\(Test)}\\
    \shline
    \multirow{2}{*}{ViViT}& \multirow{2}{*}{BMN} & 94.33\% & 33.46\% & 33.04\%\\
    ~& ~& 95.25\% & 33.91\% & 33.38\%\\
    \hline
    \multirow{3}{*}{CSN}& \multirow{3}{*}{BMN} & 94.33\% & 38.88\%  & 38.68\%\\
    ~ & ~ & 96.07\% & 39.57\% & 39.26\% \\
    ~ & ~ & 96.07\% & 41.62\% & 41.17\% \\
    \hline
    CSN & BMN+TCA & 96.07\% & 42.74\% & 42.34\% \\
    \hline
    C+S+V & Ensemble & 96.27\%  & \textbf{44.83\%} & \textbf{44.29\%} \\
    \hline
    \multicolumn{2}{c|}{2020 Winner~\cite{qing2020tfn}} & 94.33\%  & 40.55\% & 40.53\% \\
\end{tabular}\\
    \caption{\textbf{Performance comparison between Validation set and Test set on different settings.} The ``C+S+V'' in the table refers to CSN~\cite{tran2019csn}+Slowfast~\cite{feichtenhofer2019slowfast}+ViViT~\cite{arnab2021vivit}, and the ``BMN+TCA" is the candidate proposals output by BMN~\cite{lin2019bmn} are input to TCA~\cite{qing2021tca} for further refine.}
    \label{tab:test_results}
\end{table}

\section{Experiments}
The Table~\ref{tab:feature_results} explores convolution-based and transformer-based features in terms of mAP. We notice that the convolution-based methods are better than transformer-based methods for action proposals. However, as shown in the Table~\ref{tab:cls}, the transformer-based achieve impressive performance on clip-level classification task. This enlightens us that the classification performance of the backbone is not always positive for proposals, especially for Timesformer~\cite{bertasius2021timesformer}.

For LGTE~\cite{qing2021tca} in the Table~\ref{tab:resolution_results}, the improvement for CSN~\cite{tran2019csn} feature is greater than the SlowFast~\cite{feichtenhofer2019slowfast} and the ViViT~\cite{feichtenhofer2019slowfast}. For the ViViT feature, since the spatio-temporal attention has been employed, the role of LGTE is limited.

In Table~\ref{tab:resolution_results}, we explore the influence of frame resolution in extracting features. The 224x224 cropping area in training limits the spatial information of each frame, especially for non-square video frames. Therefore, adopting a larger area for cropping can improve the quality of the extracted features. This is convenient to implement for convolution-based networks. However, for the transformer-based networks with the fixed position embedding, the same resolution as the training process is still used to extract features.

In Table~\ref{tab:test_results},we show the results of our previous submissions. It can be noted that TCANet~\cite{qing2021tca} can still achieve 1.09\% improvement on the validation set, even based on a better baseline. Finally, we fuse the confidence maps output by multiple BMNs~\cite{lin2019bmn} and the refined proposals output by multiple TCANets. Thanks to the complementarity between the models trained with different features, we reached 44.83\% and 44.29\% on the validation set and test set, respectively, which was 3.76\% higher than the 2020 Winner~\cite{qing2020tfn} on the test set.

\section{Conclusion}
In this technical paper, we explore different features, resolution, BMN and TCANet for temporal action detection. We found that the features extracted by the network with high classification performance may not necessarily generate high-quality proposals. This may be guiding us to design a backbone that is more suitable for temporal action detection. The ablation studies for resolution prove that avoiding the loss of spatial information can effectively improve the performance of temporal detection. Finally, with these strategies, our single-model suppresses 1.81\% than the multi-model fusion used by the 2020 winner.

\section{Acknowledgment}
This work is supported by the National Natural Science Foundation of China under grant 61871435 and the Fundamental Research Funds for the Central Universities no. 2019kfyXKJC024 and by Alibaba Group through Alibaba Research Intern Program.

{\small
\bibliographystyle{ieee_fullname}
\bibliography{egbib}
}

\end{document}